\documentclass{article}
\usepackage{spconf,amsmath,amssymb,graphicx,cite,subfig,booktabs,url}

\def\EE{{\mathbb E}}

\DeclareMathOperator*{\argmax}{argmax}
\title{Cycle-consistency training for end-to-end speech recognition}
\name{Takaaki Hori$^1$, Ramon Astudillo$^2$, Tomoki Hayashi$^3$, Yu Zhang$^4$, Shinji Watanabe$^5$, Jonathan Le Roux$^1$}
\address{
     $^1$Mitsubishi Electric Research Laboratories (MERL), \\
     $^2$Spoken Language Systems Lab, INESC-ID,~ 
     $^3$Nagoya University,~ $^4$Google, Inc.,\\
     $^5$Center for Language and Speech Processing, Johns Hopkins University \\
{\small \tt \{thori, leroux\}@merl.com, ramon@astudillo.com, hayashi.tomoki@g.sp.m.is.nagoya-u.ac.jp,} \\
{\small \tt ngyuzh@google.com, shinjiw@jhu.edu}}
\begin{document}
\ninept
\maketitle
\begin{abstract}
This paper presents a method to train end-to-end automatic speech recognition (ASR) models using unpaired data.
Although the end-to-end approach can eliminate the need for expert knowledge such as pronunciation dictionaries to build ASR systems, it still requires a large amount of paired data, i.e., speech utterances and their transcriptions.
Cycle-consistency losses have been recently proposed as a way to mitigate the problem of limited paired data. These approaches compose a reverse operation with a given transformation, e.g., text-to-speech (TTS) with ASR, to build a loss that only requires unsupervised data, speech in this example. Applying cycle consistency to ASR models is not trivial since fundamental information, such as speaker traits, are lost in the intermediate text bottleneck. To solve this problem, this work presents a loss that is based on the speech encoder state sequence instead of the raw speech signal. This is achieved by training a Text-To-Encoder model and defining a loss based on the encoder reconstruction error.
Experimental results on the LibriSpeech corpus show that the proposed cycle-consistency training reduced the word error rate by 14.7\% from an initial model trained with 100-hour paired data, using an additional 360 hours of audio data without transcriptions.
We also investigate the use of text-only data mainly for language modeling to further improve the performance in the unpaired data training scenario.
\end{abstract}
\section{Introduction}
\label{sec:intro}

\let\thefootnote\relax\footnotetext{The work reported here was conducted at the 2018 Frederick Jelinek Memorial Summer Workshop on Speech and Language Technologies, and supported by Johns Hopkins University with unrestricted gifts from Amazon, Facebook, Google, Microsoft and Mitsubishi Electric. Work by Ram\'{o}n Astudillo was supported by the Portuguese Foundation for Science and Technology (FCT) grant number \textrm{UID/CEC/50021/2019}.}
In recent years, automatic speech recognition (ASR) technology has been widely used as an effective user interface for various devices such as car navigation systems, smart phones, and smart speakers.
The recognition accuracy has dramatically improved with the help of deep learning techniques \cite{hinton2012deep}, and reliability of speech interfaces has been greatly enhanced.
However, building ASR systems is very costly and time consuming.
Current systems typically have a module-based architecture including an acoustic model, a pronunciation dictionary, and a language model,
which rely on phonetically-designed phone units and word-level pronunciations using linguistic assumptions.
To build a language model, text preprocessing such as tokenization for some languages that do not explicitly have word boundaries is also required.
Consequently, it is not easy for non-experts to develop ASR systems, especially for underresourced languages. 

End-to-end ASR has the goal of simplifying the module-based architecture into a single-network architecture within a deep learning framework, in order to address these issues~\cite{graves2014towards, graves2013speech, chorowski2015attention, kim2016joint_icassp2017, hori-ACL-2017}.
End-to-end ASR methods typically rely only on paired acoustic and language data, without the need for extra linguistic knowledge, and train the model with a single algorithm. Therefore, this approach makes it feasible to build ASR systems without expert knowledge.
However, in the end-to-end ASR framework a large amount of training data is crucial to assure high recognition accuracy. Paired acoustic (speech) and language (transcription) realizations spoken by multiple speakers are needed~\cite{amodei2015deep}.
Nowadays, it is easy to collect audio and text data independently from the world wide web, but difficult to find paired data in different languages. Transcribing existing audio data or recording texts spoken by sufficient speakers are also very expensive.

There are several approaches that tackle the problem of limited paired data in the literature ~\cite{hayashi2018back, tjandra2017listening, Renduchintala2018, karita2018semi, Tjandra2018}. 
In particular, cycle consistency has recently been introduced in machine translation (MT)~\cite{NIPS2016_6469} and image transformation~\cite{zhu2017unpaired}, and enables one to optimize deep networks using unpaired data. The basic underlying assumption is that, given a model that converts input data to output data and another model that reconstructs the input data from the output data, input data and its reconstruction should be close to each other. For example, suppose an English-to-French MT system translates an English sentence to a French sentence, and then a French-to-English MT system back-translates the French sentence to an English sentence.
In this case, we can train the English-to-French system so that the difference between the English sentence and its back-translation becomes smaller, for which we only need English sentences.
The French-to-English MT system can also be trained in the same manner using only French sentences.

Applying the concept of cycle consistency to ASR is quite challenging. As is the case in MT, the output of ASR is a discrete distribution over the set of all possible sentences. It is therefore not possible to build an end-to-end differentiable loss that back-propagates error through the most probable sentence in this step. Since the set of possible sentences is exponentially large in the size of the sentence, it is not possible to exactly average over all possible sentences either. Furthermore, unlike in MT and image transformation, in ASR, the input and output domains are very different and do not contain the same information. The output text does not include speaker and prosody information, which is eliminated through feature extraction and decoding. Hence, the speech reconstructed by the TTS system does not have the original speaker and prosody information and can result in a strong mismatch.

Previous approaches related to cycle consistency in end-to-end ASR \cite{tjandra2017listening,Tjandra2018} circumvent these problems by avoiding back-propagating the error beyond the discrete steps and adding a speaker network to transfer the information not present in the text. Therefore, these methods are not strictly cycle-consistency training, as used in MT and image transformation. Gradients are not cycled both through ASR and TTS simultaneously and only the second step on a ASR-TTS or TTS-ASR chain can be updated. 

In this work, we propose an alternative approach that uses an end-to-end differentiable loss in the cycle-consistency manner. This idea rests on the two following principles.
\begin{enumerate}
    \item Encoder-state-level cycle consistency:\\
     We use ASR encoder state sequences for computing the cycle consistency instead of waveform or spectral features. This uses a normal TTS Tacotron2 end-to-end model \cite{shen2017natural} modified to reconstruct the encoder state sequence instead of speech. We call this a {\it text-to-encoder (TTE) model}~\cite{hayashi2018back}, which we introduced in our prior work on data augmentation.
     This approach reduces the mismatch between the original and the reconstruction by avoiding the problem of missing para-linguistic information. 
    \item Expected end-to-end loss:\\
     We use an expected loss approximated with a sampling-based method. In other words, we sample multiple sentences from the ASR model, generate an encoder state sequence for each, and compute the consistency loss for each sentence by comparing each encoder state sequence with the original. Then, the mean loss can be used to backpropagate the error to the ASR model via the REINFORCE algorithm~\cite{williams1992simple}. This allows us to update the ASR system when the TTE is used to compute the loss, unlike \cite{tjandra2017listening}. 
\end{enumerate}
The proposed approach allows therefore training with unpaired data, even if only speech is available. Furthermore, since error is backpropagated into the ASR system from a TTS-based loss, additional unsupervised losses can be used, such as language models. We demonstrate the efficacy of the proposed method in a semi-supervised training condition on the LibriSpeech corpus.

\section{Cycle-consistency training for ASR}
\label{sec:proposed}
\subsection{Basic concept}
\label{ssec:overview}
\begin{figure}[t]
\begin{center}
\includegraphics[width=1.0\columnwidth]{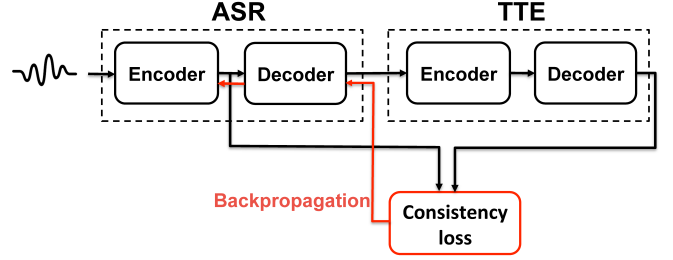}
\end{center}
\vspace{-5mm}
\caption{Cycle-consistency training for ASR.}
\label{fig:cycle_consistency}
\vspace{-3mm}
\end{figure}

The proposed method consists of an ASR encoder-decoder, a TTE encoder-decoder, and consistency loss computation as shown in Fig.~\ref{fig:cycle_consistency}. In this framework, we need only audio data for backpropagation.
In a first step, the ASR system transcribes the input audio feature sequence into a sequence of characters. In addition to this, a encoder state sequence is obtained. In a second step, the TTE system reconstructs the ASR encoder state sequence fro the character sequence. Finally, the cycle-consistency loss is computed by comparing the original state sequence and the reconstructed one. Backpropagation is performed with respect to this loss to update the ASR parameters. 

\subsection{Attention-based ASR model}
\label{ssec:att_e2e_asr}
The ASR model used is the well known attention-based encoder-decoder \cite{chan2015listen}. This model directly estimates the posterior $p_\mathrm{asr}(\mathbf{C}|\mathbf{X})$, where $\mathbf{X} = \{\mathbf{x}_1,\mathbf{x}_2, \dots, \mathbf{x}_T|\mathbf{x}_t \in \mathbb{R}^D\}$ is a sequence of input $D$-dimensional feature vectors, and $\mathbf{C} = \{c_1, c_2, \dots, c_L|c_l \in {\cal U}\}$ is a sequence of output characters in the label set ${\cal U}$.
The posterior $p_\mathrm{asr}(\mathbf{C}|\mathbf{X})$ is factorized the probability chain rule as follows:
\begin{equation}
    p_\mathrm{asr}(\mathbf{C}|\mathbf{X}) = \prod_{l=1}^L p_\mathrm{asr}(c_l | c_{1:l-1}, \mathbf{X}),
\end{equation}
where $c_{1:l-1}$ represents the subsequence $\{c_1, c_2, \dots\, c_{l-1}\}$, and\\
$p_\mathrm{asr}(c_l | c_{1:l-1}, \mathbf{X})$ is calculated as follows:
\begin{eqnarray}
    \label{eq:att_encoder}
    \mathbf{h}_t^{\mathrm{asr}} &=& \mathrm{Encoder}^{\mathrm{asr}}(\mathbf{X}), \\
    \label{eq:att_weight1}
    a_{lt}^{\mathrm{asr}} &=& \mathrm{Attention}^{\mathrm{asr}}(\mathbf{q}_{l-1}^{\mathrm{asr}},
                                                  \mathbf{h}_t^{\mathrm{asr}},
                                                  \mathbf{a}_{l-1}^{\mathrm{asr}}),\\
    \label{eq:att_weight2}
    \mathbf{r}_l^{\mathrm{asr}} &=& \Sigma_{t=1}^{T} a_{lt}^{\mathrm{asr}}\mathbf{h}_t^{\mathrm{asr}}, \\
    \label{eq:att_decoder}
    \mathbf{q}_l^{\mathrm{asr}} &=& \mathrm{Decoder}^{\mathrm{asr}}(\mathbf{r}_l^{\mathrm{asr}},
                                                                     \mathbf{q}_{l-1}^{\mathrm{asr}},
                                                                     c_{l-1}), \\
    \label{eq:att_posterior}
    p_\mathrm{asr}(c_l | c_{1:l-1}, \mathbf{X}) &=& \mathrm{Softmax}(\mathrm{LinB}(\mathbf{q}_l^{\mathrm{asr}})),
\end{eqnarray}
where $a_{lt}^{\mathrm{asr}}$ represents an attention weight, $\mathbf{a}_l^{\mathrm{asr}}$ the corresponding attention weight vector, $\mathbf{h}_t^{\mathrm{asr}}$ and $\mathbf{q}_l^{\mathrm{asr}}$ the hidden states of the encoder and decoder networks, respectively, $\mathbf{r}_l^{\mathrm{asr}}$ a character-wise hidden vector, which is a weighted summarization of the hidden vectors $\mathbf{h}_t^{\mathrm{asr}}$ using the attention weight vector $\mathbf{a}_l^{\mathrm{asr}}$, and $\mathrm{LinB}(\cdot)$ represents a linear layer with a trainable matrix and bias parameters.

All of the above networks are optimized using back-propagation to minimize the following objective function:
\begin{equation}
\begin{split}
    \label{eq:att_obj}
    \mathcal{L}_\mathrm{asr} & = - \log p_\mathrm{asr}(\mathbf{C} | \mathbf{X}) \\
                & = - \Sigma_{l=1}^L \log p_\mathrm{asr}(c_l^{\mathrm{asr}} | c_{1:l-1}^{\mathrm{asr}}, \mathbf{X}),
\end{split}
\end{equation}
where $c_{1:l-1}^{\mathrm{asr}} = \{c_1^{\mathrm{asr}}, c_2^{\mathrm{asr}}, \dots, c_{l-1}^{\mathrm{asr}}\}$ represents the ground truth for the previous characters, i.e. teacher-forcing is used in training. In the inference stage, the character sequence $\hat{\bf C}$ is predicted as
\begin{align}
    \hat{\mathbf{C}}=\argmax_{\mathbf{C} \in {\cal U}^+} \log p_\mathrm{asr}(\mathbf{C}|\mathbf{X}).
    \label{eq:asr_argmax}
\end{align}
where ${\cal U}^+$ is the set of all sentences formed from the original character vocabulary ${\cal U}$.

\subsection{Tacotron2-based TTE model}
\label{ssec:e2e_tts}
For the TTE model, we use the Tacotron2 architecture, which has demonstrated superior performance in the field of text-to-speech synthesis~\cite{shen2017natural}.
In our framework, the network predicts the ASR encoder state $\mathbf{h}_t^{\mathrm{asr}}$ and the end-of-sequence probability $s_t$ at each frame $t$ from a sequence of input characters $\mathbf{C} = \{c_1, c_2, \dots, c_L\}$ as follows:{\allowdisplaybreaks
\begin{eqnarray}
    \label{eq:tts_encoder}
    \mathbf{h}_l^{\mathrm{tte}} &=& \mathrm{Encoder}^{\mathrm{tte}}(\mathbf{C}), \\
    \label{eq:tts_attention}
    a_{tl}^{\mathrm{tte}} &=& \mathrm{Attention}^{\mathrm{tte}}(\mathbf{q}_{t-1}^{\mathrm{tte}},
                                                  \mathbf{h}_l^{\mathrm{tte}},
                                                  \mathbf{a}_{t-1}^{\mathrm{tte}}),\\
    \label{eq:tts_attention2}
    \mathbf{r}_t^{\mathrm{tte}} &=& \Sigma_{l=1}^{L} a_{tl}^{\mathrm{tte}}\mathbf{h}_l^{\mathrm{tte}}, \\
    \mathbf{v}_{t-1} &=& \mathrm{Prenet}(\mathbf{h}_{t-1}^{\mathrm{asr}}), \\
    \label{eq:tts_decoder}
    \mathbf{q}_t^{\mathrm{tte}} &=& \mathrm{Decoder}^{\mathrm{tte}}(\mathbf{r}_t^{\mathrm{tte}}, \mathbf{q}_{t-1}^{\mathrm{tte}}, \mathbf{v}_{t-1}), \\
    \label{eq:tts_output_before}
    \hat{\mathbf{h}}_t^{b, \mathrm{asr}} &=& \tanh(\mathrm{LinB}(\mathbf{q}_t^{\mathrm{tte}})), \\
    \mathbf{d}_{t} &=& \mathrm{Postnet}(\mathbf{q}_l^{\mathrm{tte}}), \\
	\label{eq:tts_output_after}
    \hat{\mathbf{h}}_t^{a,\mathrm{asr}} &=& \tanh(\mathrm{LinB}(\mathbf{q}_l^{\mathrm{tte}})+ \mathbf{d}_{t}), \\
    \hat{s}_t &=& \mathrm{Sigmoid}(\mathrm{LinB}(\mathbf{q}_t^{\mathrm{tte}})),
\end{eqnarray}}
where $\mathrm{Prenet}(\cdot)$ is a shallow feed-forward network to convert the network outputs before feedback to the decoder, $\mathrm{Postnet}(\cdot)$ is a convolutional neural network to refine the network outputs, and $\hat{\mathbf{h}}_t^{b,\mathrm{asr}}$ and $\hat{\mathbf{h}}_t^{a,\mathrm{asr}}$ represent predicted hidden states of the ASR encoder before and after refinement by Postnet.
Note that the indices $t$ and $l$ of the encoder and decoder states are reversed compared to the ASR formulation in Eqs.~(\ref{eq:att_encoder})-(\ref{eq:att_posterior}), and that we use an additional activation function $\tanh(\cdot)$ in Eqs.~(\ref{eq:tts_output_before}) and (\ref{eq:tts_output_after}) to avoid range mismatch in the outputs, in contrast to the original Tacotron2~\cite{shen2017natural}.

All of the networks are jointly optimized to minimize the following objective function:
\begin{multline}
    \label{eq:tts_obj}
    \mathcal{L}_\mathrm{tte} = \mathrm{MSE}(\hat{\mathbf{h}}_t^{a,\mathrm{asr}}, \mathbf{h}_t^{\mathrm{asr}}) + \mathrm{MSE}(\hat{\mathbf{h}}_t^{b,\mathrm{asr}}, \mathbf{h}_t^{\mathrm{asr}})\\
    + \mathrm{L1}(\hat{\mathbf{h}}_t^{a,\mathrm{asr}}, \mathbf{h}_t^{\mathrm{asr}}) + \mathrm{L1}(\hat{\mathbf{h}}_t^{b,\mathrm{asr}}, \mathbf{h}_t^{\mathrm{asr}})\\
    + \tfrac{1}{T}\Sigma_{t=1}^T (s_t \ln \hat{s}_t + (1 - s_t) \ln (1 - \hat{s}_t)),
\end{multline}
where $\mathrm{MSE}(\cdot)$ represents mean square error, $\mathrm{L1}(\cdot)$ represent an L1 norm, and the last two terms represent the binary cross entropy for the end-of-sequence probability.

\subsection{Cycle-consistency training}
In this work, we use the TTE reconstruction loss ${\cal L}_\mathrm{tte}$ in Eq.~\eqref{eq:tts_obj} to measure the cycle consistency. The loss compares the ASR encoder state sequence with the encoder sequence reconstructed from the ASR output by the TTE.
However, the argmax function in Eq.~\eqref{eq:asr_argmax} to output the character sequence is not differentiable, and the consistency loss cannot be propagated through TTE to ASR directly. To solve this problem, we introduce the expected loss 
\begin{align}
\label{eq:expected_tts_obj}
{\cal L}_\mathrm{ette} & =\EE_{\mathbf{C}|\mathbf{X}}\left[{\cal L}_\mathrm{tte}(\hat{\mathbf{H}}^\mathrm{asr}(\mathbf{C}), \mathbf{H}^\mathrm{asr}(\mathbf{X})) \right], 
\end{align}
where $\hat{\mathbf{H}}^\mathrm{asr}(\mathbf{C})$ denotes the state sequence $\{\hat{\mathbf{h}}_t^{a,\mathrm{asr}}, \hat{\mathbf{h}}_t^{b,\mathrm{asr}}, \hat{s}_t|t=1,\dots,T\}$ predicted by the TTE model for a given character sequence $\mathbf{C}$, and $\mathbf{H}^\mathrm{asr}(\mathbf{X})$ denotes the original state sequence $\{\mathbf{h}_t^\mathrm{asr}, s_t|t=1,\dots,T\}$ given by the ASR encoder for the input feature sequence $\bf X$.

To compute the gradients with respect to the expectation in Eq.~\ref{eq:expected_tts_obj}, we utilize the REINFORCE algorithm~\cite{williams1992simple}. This yields the following expression for the gradient 
\begin{equation}
\nabla {\cal L}_\mathrm{ette} \approx \frac{1}{N}\sum_{\substack{{\bf C}^n\sim p_\mathrm{asr}(\cdot|\mathbf{X}),\\ n=1,\dots, N}} \!\!\! \mathrm{T}\left(\mathbf{C}^n, \mathbf{X}\right) 
          \nabla \log p_\mathrm{asr}(\mathbf{C}^n|\mathbf{X}),
\end{equation}
\noindent where the weight for each sample $\mathbf{C}^n$ is defined as 
\begin{equation}
\mathrm{T}\left(\mathbf{C}^n, \mathbf{X}\right) = {\cal L}_\mathrm{tte}(\hat{\mathbf{H}}^\mathrm{asr}(\mathbf{C}^n), \mathbf{H}^\mathrm{asr}(\mathbf{X})) - B(\mathbf{X}, \mathbf{C}^n)
\end{equation}
and $B({\bf X},{\bf C}^n)$ is a baseline value used to reduce the estimate variance~\cite{williams1992simple}. We used the mean value of $\mathbf{H}^\mathrm{asr}(\mathbf{C}^n)$ over $N$ samples for $B({\bf X},{\bf C}^n)$ in this work.

\section{Related work}
The algorithm introduced in this paper is related to existing works on data augmentation and chain-based training. Our prior work ~\cite{hayashi2018back} introduced the TTE model but used the synthesized encoder state sequences to train the ASR decoder from text data only. This is equivalent to back-translation in MT \cite{sennrich2015improving} and builds a non-differentiable TTE-ASR chain as opposed to the end-to-end differentiable ASR-TTE chain proposed here.

The work in \cite{karita2018semi} introduces a model consisting of a text-to-text auto-encoder and a speech-to-text encoder-decoder sharing the speech and text encodings. This model can also be trained jointly using paired and unpaired data but uses a simpler text encoder. Furthermore speech-only data is used to enhance the speech encodings, but not used to reduce recognition errors unlike our cycle-consistency approach. Finally, the text encoder is much simpler than our TTE model. In our work, the TTE model can hopefully generate better speech encodings to compute the consistency loss.

The speech chain model~\cite{tjandra2017listening} is the most similar architecture to ours. As described in Section \ref{sec:intro}, the ASR model is trained with synthesized speech and the TTS model is trained with ASR hypotheses for unpaired data. Therefore, the models are not tightly connected with each other, i.e., one model cannot be updated directly with the help of the other model to reduce the recognition or synthesis errors. Our approach utilizes an end-to-end differentiable loss that allows TTS or other loss to be used \textit{after} ASR for unsupervised training. We introduce as well the TTE model, which benefits from the reduction of speaker variations in the loss function and of computational complexity. With regard to cycle-consistency approaches in other disciplines, our approach is most similar to the dual learning approach in MT \cite{NIPS2016_6469}. This paper combines alternating losses as in ~\cite{tjandra2017listening} using REINFORCE to compute expected translation losses.  
\section{EXPERIMENTS}
\label{sec:experiment}
\subsection{Conditions}
We conducted several experiments using the LibriSpeech corpus~\cite{panayotov2015librispeech}, consisting of two sets of clean speech data (100 hours + 360 hours), and other (noisy) speech data (500 hours) for training.
We used 100 hours of the clean speech data to train the initial ASR and TTE models, and the audio of 360 hours set for unsupervised re-training of the ASR model with the cycle-consistency loss.
We used five hours of clean development data as a validation set, and five hours of clean test data as an evaluation set.

The open source speech recognition toolkit Kaldi~\cite{Povey_ASRU2011} was used to extract 80-dimensional log mel-filter bank acoustic vectors with three-dimensional pitch features.
The ASR encoder had an eight-layered bidirectional long short-term memory with 320 cells including projection layers~\cite{sak2014long} (BLSTMP), and the ASR decoder had a one-layered LSTM with 300 cells.
In the second and third layers from the bottom of the ASR encoder, sub-sampling was performed to reduce the utterance length from $T$ down to $T/4$.
The ASR attention network used location-aware attention~\cite{chorowski2015attention}.
For decoding, we used a beam search algorithm with beam size of 20.
We set the maximum and minimum lengths of the output sequence to 0.2 and 0.8 times the length of the subsampled input sequence, respectively.

The architecture of the TTE model followed the original Tacotron2~\cite{shen2017natural}.
It use 512-dimensional character embeddings,
the TTE encoder consisted of a three-layered 1D convolutional neural network (CNN) containing 512 filters with size 5, a batch normalization, and rectified linear unit (ReLU) activation function, and a one-layered BLSTM with 512 units (256 units for forward processing, the rest for backward processing).
Although the attention mechanism of the TTE model was based on location-aware attention~\cite{chorowski2015attention}, we additionally accumulated the attention weight feedback to the next step to accelerate attention learning.
The TTE decoder consisted of a two-layered LSTM with 1024 units.
Prenet was a two-layered feed forward network with 256 units and ReLU activation.
Postnet was a five-layered CNN containing 512 filters with the shape 5, a batch normalization, and tanh activation function except in the final layer.
Dropout~\cite{srivastava2014dropout} with a probability of 0.5 was applied to all of the  convolution and Prenet layers.
Zoneout~\cite{krueger2016zoneout} with a probability of 0.1 was applied to the decoder LSTM.
During generation, we applied dropout to Prenet in the same manner as in~\cite{shen2017natural}, and set the threshold value of the end-of-sequence probability at 0.75 to prevent from cutting off the end of the input sequence.

In cycle-consistency training, five sequences of characters were drawn from the ASR model for each utterance, where each character was drawn repeatedly from the Softmax distribution of ASR until it encountered the end-of-sequence label `\texttt{<eos>}'.
During training, we also used the 100-hour paired data to regularize the model parameters in a teacher-forcing manner, i.e., the parameters were updated alternately by cross-entropy loss with paired data and the cycle-consistency loss with unpaired data.

All models were trained using the end-to-end speech processing toolkit ESPnet~\cite{watanabe2018espnet} on a single GPU (Titan Xp).
Character error rate (CER) and word error rate (WER) were used as evaluation metrics.

\begin{figure}[t]
\begin{center}
\includegraphics[width=1.0\columnwidth]{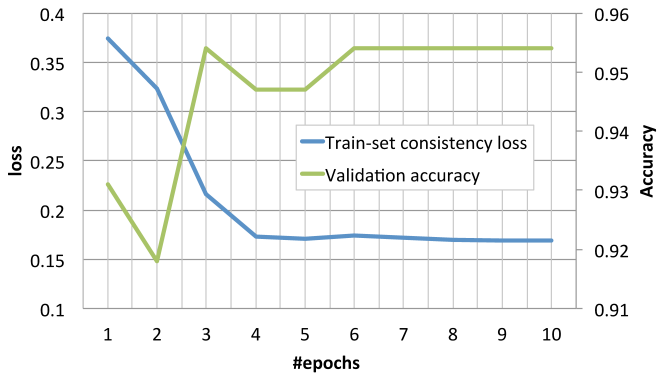}
\end{center}
\vspace{-5mm}
\caption{Learning curve.}
\label{fig:learning_curve}
\vspace{-2mm}
\end{figure}
\subsection{Results}
First, we show the changes of the consistency loss for training data and the validation accuracy for development data in Fig.~\ref{fig:learning_curve},
where the accuracy was computed based on the prediction with ground truth history.
The consistency loss successfully decreased as the number of epochs increased. Although the validation accuracy did not improve smoothly, it reached a better value than that for the first epoch. We chose the 6th-epoch model for the following ASR experiments.

Table~\ref{tb:asr_results} shows the ASR performance using different training methods.
Compared with the baseline result given by the initial ASR model, we can confirm that our proposed cycle-consistency training reduced the word error rate from 25.2\% to 21.5\%, a relative reduction of 14.7\%\footnote{Our baseline WER is much worse than that reported in \cite{panayotov2015librispeech} for the 100-hour training setup. This is because we did not use any pronunciation lexicon or word-based language model for end-to-end ASR. Such end-to-end systems typically underperform conventional DNN/HMM systems with n-gram language model when using this size of training data.}.
Thus, the results demonstrate that the proposed method works for ASR training with unpaired data.
To verify the effectiveness of our approach, we further examined more straightforward methods, in which we simply used cross-entropy (CE) loss for unpaired data, where the target was chosen as the one best ASR hypothesis or sampled in the same manner as the cycle-consistency training. To alleviate the impact of the ASR errors, we weighted the CE loss by 0.1 for unpaired data while we did not down-weight the paired data. However, the error rates increased significantly in the 1-best condition. Even in the 5-sample condition, we could not obtain better performance than the baseline.
We also conducted additional experiments under an oracle condition, where the 360-hour paired data were used together with the 100-hour data using the standard CE loss. The error rates can be considered the upper bound of this framework. We can see that there is still a big gap to the upper bound and further challenges need to be overcome to reach this goal.

Finally, we combined the ASR model with a character-based language model (LM) in a shallow fusion technique~\cite{hori2017advances}. An LSTM-based LM was trained using text-only data from the 500-hour noisy set excluding audio data, and used for decoding.
As shown in Table~\ref{tb:asr_results_lm}, the use of  
text-only data yielded further improvement reaching 19.5\% WER (an 8\% error reduction), which is the best number we have achieved so far for this unpaired data setup.

\begin{table}[t]
\begin{center}
\caption{ASR performance using different training methods.}
\label{tb:asr_results}
\scalebox{0.99}{%
{
\begin{tabular}{l c c c}
                                & \multicolumn{2}{c}{CER / WER [\%]} \\ \cmidrule{2-3}
                                & Validation   & Evaluation  \\ \toprule
Baseline                        & 11.2 / 24.9  & 11.1 / 25.2 \\
Cycle-consistency loss	        & {\bf \phantom{1}9.5} / {\bf 21.5} & {\bf \phantom{1}9.4} / {\bf 21.5} \\ \midrule
CE loss (1 best)	            & 47.8 / 86.8 &	48.8 / 89.3 \\
CE loss (5 samples)             & 13.3 / 28.2 & 12.3 / 27.7 \\
 \midrule
Oracle                          & \phantom{1}4.7 / 11.4   & \phantom{1}4.6 / 11.8  \\
 \bottomrule
\end{tabular}
}}
\end{center}
\end{table}
\vspace{-2mm}
\begin{table}[t]
\begin{center}
\caption{ASR performance with LM shallow fusion.}
\label{tb:asr_results_lm}
\scalebox{0.99}{%
{
\begin{tabular}{l c c c}
                      & \multicolumn{2}{c}{CER / WER [\%]} \\ \cmidrule{2-3}
                      & Validation   & Evaluation  \\ \toprule
Baseline + LM	      & 11.9 / 22.6  & 11.9 / 22.9 \\
Cycle consistency + LM & {\bf 10.2} / {\bf 19.6} & {\bf \phantom{1}9.9} / {\bf 19.5} \\ \bottomrule
\end{tabular}
}}
\vspace{-4mm}
\end{center}
\end{table}

\section{CONCLUSION}
\label{sec:conclusion}
In this paper, we proposed a novel method to train end-to-end automatic speech recognition (ASR) models using unpaired data.
The method employs an attention-based ASR model and a Tacotron2-based text-to-encoder (TTE) model to compute a cycle-consistency loss using audio data only.
Experimental results on the LibriSpeech corpus demonstrated that the proposed cycle-consistency training reduced the word error rate by 14.7\% from an initial model trained with 100-hour paired data, using an additional 360 hours of audio-only data without transcriptions.
We also investigated the use of text-only data from 500-hour utterances for language modeling, and obtained a further error reduction of 8\%.
Accordingly, we achieved 22.7\% error reduction in total for this unpaired data setup.
Future work includes joint training of ASR and TTE model using both sides of the cycle-consistency loss, and the use of additional loss functions to make the training better.

\bibliographystyle{IEEEbib}
\bibliography{strings,refs}

\end{document}